\documentclass[conference]{IEEEtran}
\IEEEoverridecommandlockouts

\usepackage{cite}
\usepackage{amsmath,amssymb,amsfonts}
\usepackage{algorithmic}
\usepackage{graphicx}
\usepackage{textcomp}
\usepackage{xcolor}
\def\BibTeX{{\rm B\kern-.05em{\sc i\kern-.025em b}\kern-.08em
    T\kern-.1667em\lower.7ex\hbox{E}\kern-.125emX}}
\begin{document}

\title{SubstationAI: Multimodal Large Model-Based Approaches for Analyzing Substation Equipment Faults\\
\thanks{*Corresponding author. qkpeng@xjtu.edu.cn}
}

\author{
\IEEEauthorblockN{1\textsuperscript{st} Jinzhi Wang}
\IEEEauthorblockA{\textit{Systems Engineering Institute} \\
\textit{Xi'an Jiaotong University}\\
Xian, China\\
wangjz5515@stu.xjtu.edu.cn}
\and
\IEEEauthorblockN{2\textsuperscript{nd} Qinfeng Song}
\IEEEauthorblockA{\textit{Substation Maintenance Center} \\
\textit{Hefei Electric Power Supply Company}\\
Hefei, China \\
songqingfeng1166@sina.com}
\and
\IEEEauthorblockN{3\textsuperscript{rd}Lidong Qian}
\IEEEauthorblockA{\textit{Substation Maintenance Center} \\
\textit{Hefei Electric Power Supply Company}\\
Hefei, China\\
lidongqianah@sina.com}
\and
    \IEEEauthorblockN{4\textsuperscript{th} Haozhou Li}
\IEEEauthorblockA{\textit{Systems Engineering Institute} \\
\textit{Xi'an Jiaotong University}\\
Xian, China\\
lihaozhou1126@stu.xjtu.edu.cn}
\and
\IEEEauthorblockN{5\textsuperscript{th} Qinke Peng\textsuperscript{*}}
\IEEEauthorblockA{\textit{Systems Engineering Institute} \\
\textit{Xi'an Jiaotong University}\\
Xian, China\\
qkpeng@xjtu.edu.cn}
\and
\IEEEauthorblockN{6\textsuperscript{th} Jiangbo Zhang}
\IEEEauthorblockA{\textit{School of Electronic Science and Technology} \\
\textit{Xi'an Jiaotong University}\\
Xian, China\\
4522253009@stu.xjtu.edu.cn}
}
\maketitle

\begin{abstract}
The reliability of substation equipment is crucial to the stability of power systems, but traditional fault analysis methods heavily rely on manual expertise, limiting their effectiveness in handling complex and large-scale data. This paper proposes a substation equipment fault analysis method based on a multimodal large language model (MLLM). We developed a database containing 40,000 entries, including images, defect labels, and analysis reports, and used an image-to-video generation model for data augmentation. Detailed fault analysis reports were generated using GPT-4. Based on this database, we developed SubstationAI, the first model dedicated to substation fault analysis, and designed a fault diagnosis knowledge base along with knowledge enhancement methods. Experimental results show that SubstationAI significantly outperforms existing models, such as GPT-4, across various evaluation metrics, demonstrating higher accuracy and practicality in fault cause analysis, repair suggestions, and preventive measures, providing a more advanced solution for substation equipment fault analysis.
\end{abstract}


\section{Introduction}
Existing research on substation fault analysis primarily relies on traditional manual expertise and simple detection models\cite{b1}. However, these methods face significant challenges when dealing with complex equipment failures\cite{b2}. Substation equipment faults often involve various intricate factors, such as transformer damage and cable insulation failure, which traditional methods struggle to comprehensively capture and analyze. Additionally, current fault detection systems often focus solely on preliminary identification and fail to delve into the root causes of faults or provide effective repair recommendations, making fault analysis more challenging and time-consuming\cite{b3}. Therefore, it is essential to incorporate advanced multimodal large model (MLLM) technologies to develop intelligent fault analysis systems. Such systems would not only enhance fault detection accuracy but also provide in-depth analysis of fault causes and targeted repair recommendations, significantly improving the management and maintenance efficiency of substation equipment.

In recent years, multi-modal large models (e.g., GPT-4\cite{b4}) have gained attention for their outstanding ability to process and integrate various types of data, such as text\cite{b5}\cite{b6}, images\cite{b7}\cite{b8}, and time series\cite{b9}. These models can synthesize information from different data sources to offer more comprehensive solutions in complex application scenarios. For instance, GPT-4 has demonstrated exceptional performance in multi-modal understanding and generation. Compared to traditional single-modal models, it exhibits superior generalization and precision when handling multiple data inputs. This capability has enabled multimodal large models to excel in tasks such as image generation\cite{b10}\cite{b11}, cross-modal retrieval\cite{b12}\cite{b13}, and complex reasoning\cite{b14}\cite{b15}, presenting new opportunities for professional fields like equipment fault analysis. By combining different data types, these models can achieve more precise and comprehensive analysis and decision support.

While multi-modal large models excel in handling diverse data types, they often underperform in specialized domains due to a lack of domain-specific knowledge\cite{b16}, particularly in substation equipment fault analysis. These models struggle to interpret fault images, identify fault types, analyze root causes, and provide improvement suggestions.

This paper introduces SubstationAI, a multi-modal large language model specifically designed for substation fault analysis. Utilizing the substation defect detection dataset from the China Electric Power Research Institute and the State Grid Corporation of China’s guidelines, we developed detailed fault analysis reports and integrated a specialized knowledge base, expanding the dataset to 40,000 samples. Our model is fine-tuned on the LLAVA1.5-7B model, enhanced with relevant knowledge for this domain.

In summary, our contributions are threefold: 
\begin{itemize}
    \item We constructed the first substation fault analysis dataset and established four evaluation metrics to assess the quality of the analysis reports generated by the LLM.
    \item We proposed a knowledge enhancement method for substation fault analysis, which significantly improves performance metrics across different model scales.
    \item We trained SubstationAI—the first multi-modal model specifically for substation equipment fault analysis—and achieved superior scores compared to other models in this domain.
\end{itemize}

\section{Task Setup}
This task generates detailed fault analysis reports from substation fault images. The input consists of fault images and prompts that guide the model in report generation. The output is a fault analysis report that identifies the fault type, analyzes the causes, and provides repair recommendations. The mathematical formula is as follows:
\[
R = \text{GenReport}\left( \text{Fuse}\left( \text{ImgFeat}(I), \text{TxtEnc}(P) \right) \right)
= \left( T, C, S \right)
\]
In this formula,Here, \( R \) represents the generated report. \(\text{ImgFeat}(I)\) denotes the image feature extractor applied to the fault image \(I\), and \(\text{TxtEnc}(P)\) refers to the text encoder used for processing the prompt \(P\). The function \(\text{Fuse}\) is the fusion function that combines the image features and text encodings. \(\text{GenReport}\) is the report generation function. The generated report \(R\) consists of three main components: \( T \) indicates the fault type, \( C \) represents the fault cause analysis, and \( S \) provides the repair suggestions.

\subsection{Evaluation metrics}\label{AA}
Existing metrics such as BLEU and ROUGE are typically used for tasks like translation and summarization. However, due to the specific context of substation fault analysis, these metrics do not fully address the evaluation needs for substation fault analysis reports. Based on discussions with electrical power experts and considering the practical requirements of substation fault analysis, as well as referencing metrics from fields such as healthcare\cite{b17}\cite{b18}, education\cite{b19}, and software engineering\cite{b20}, we have identified four key characteristics that a high-quality substation fault analysis report should exhibit:

\begin{itemize}
    \item Accuracy: The report should accurately identify the fault type and cause, providing detailed descriptions of the specific fault, potential reasons, and impact. Vague fault descriptions or lack of technical details can lead to misleading conclusions, indicating a lack of professionalism.
    \item Clarity: The report should convey fault information in clear and concise language, systematically presenting fault descriptions, impact scope, and repair steps. Overly complex information or obscure terminology can confuse users and reduce the report’s effectiveness.
    \item Completeness: The report should comprehensively consider all possible fault causes and provide detailed analysis and repair suggestions. If only partial causes are listed or repair advice is omitted, it may fail to offer an effective solution.
    \item Practicality: The report should provide feasible repair suggestions and clear steps to effectively guide maintenance work. Abstract or vague recommendations may complicate the repair process and reduce the report’s practicality.
\end{itemize}

These four metrics are crucial for evaluating substation fault analysis reports. Only reports that excel in these areas can significantly enhance the accuracy of fault diagnosis and the efficiency of equipment maintenance. Based on these metrics, we have established the following scoring standards, defining a rating scale from 1 to 5, from low to high, to ensure the independence and objectivity of the evaluation process.

\section{Datasets}
\subsection{Dataset Source}
The dataset utilized in this study is primarily derived from the publicly available substation defect detection dataset provided by the China Electric Power Research Institute\cite{b21}. This dataset includes 10,330 substation fault images, annotated with 14 different types of substation faults and their corresponding fault location labels.
\begin{figure}
    \centering
    \includegraphics[width=1\linewidth]{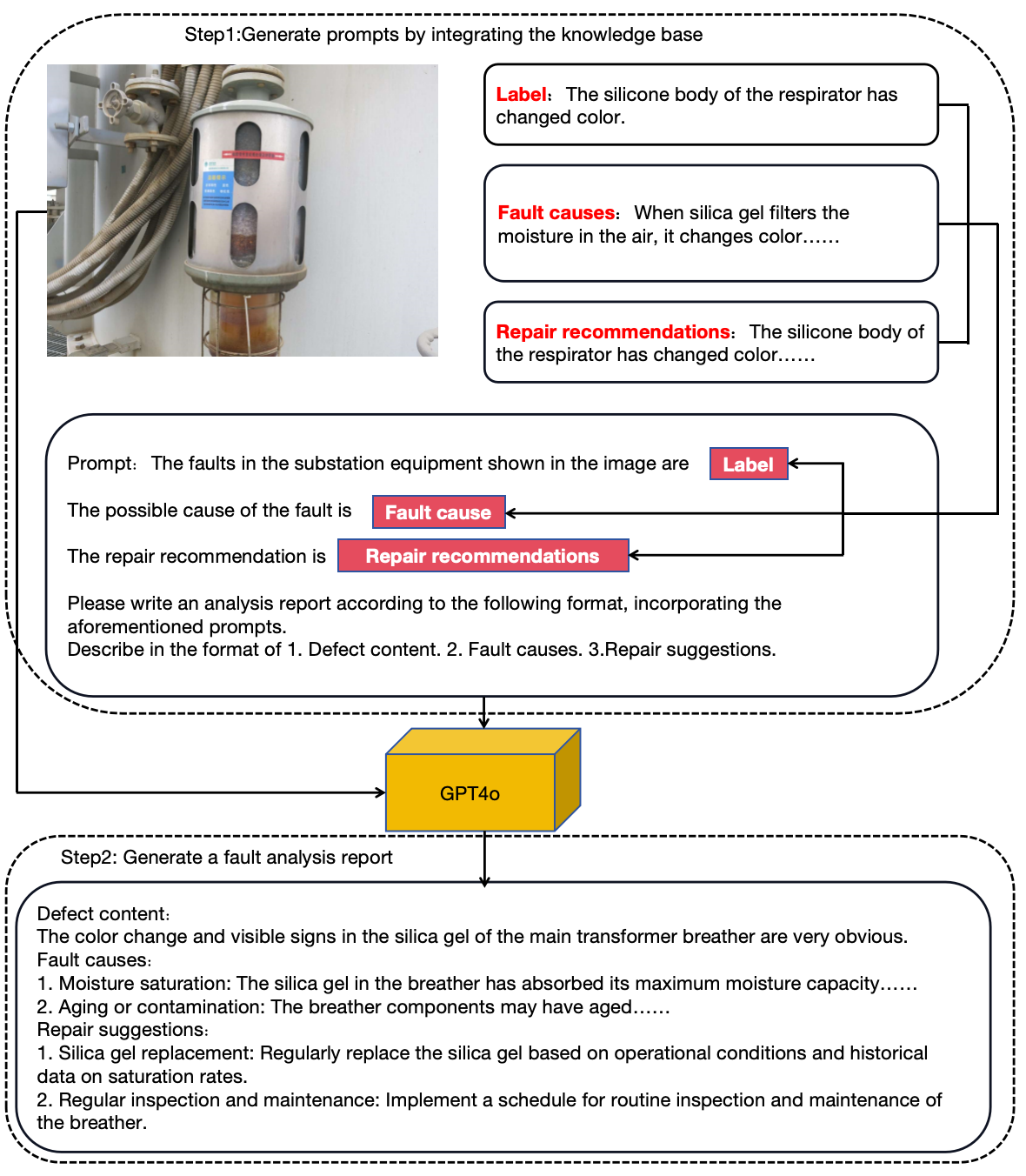}
    \caption{Database Construction Diagram}
    \label{fig1}
\end{figure}
\subsection{Dataset Construction}
\begin{figure}
    \centering
    \includegraphics[width=1\linewidth]{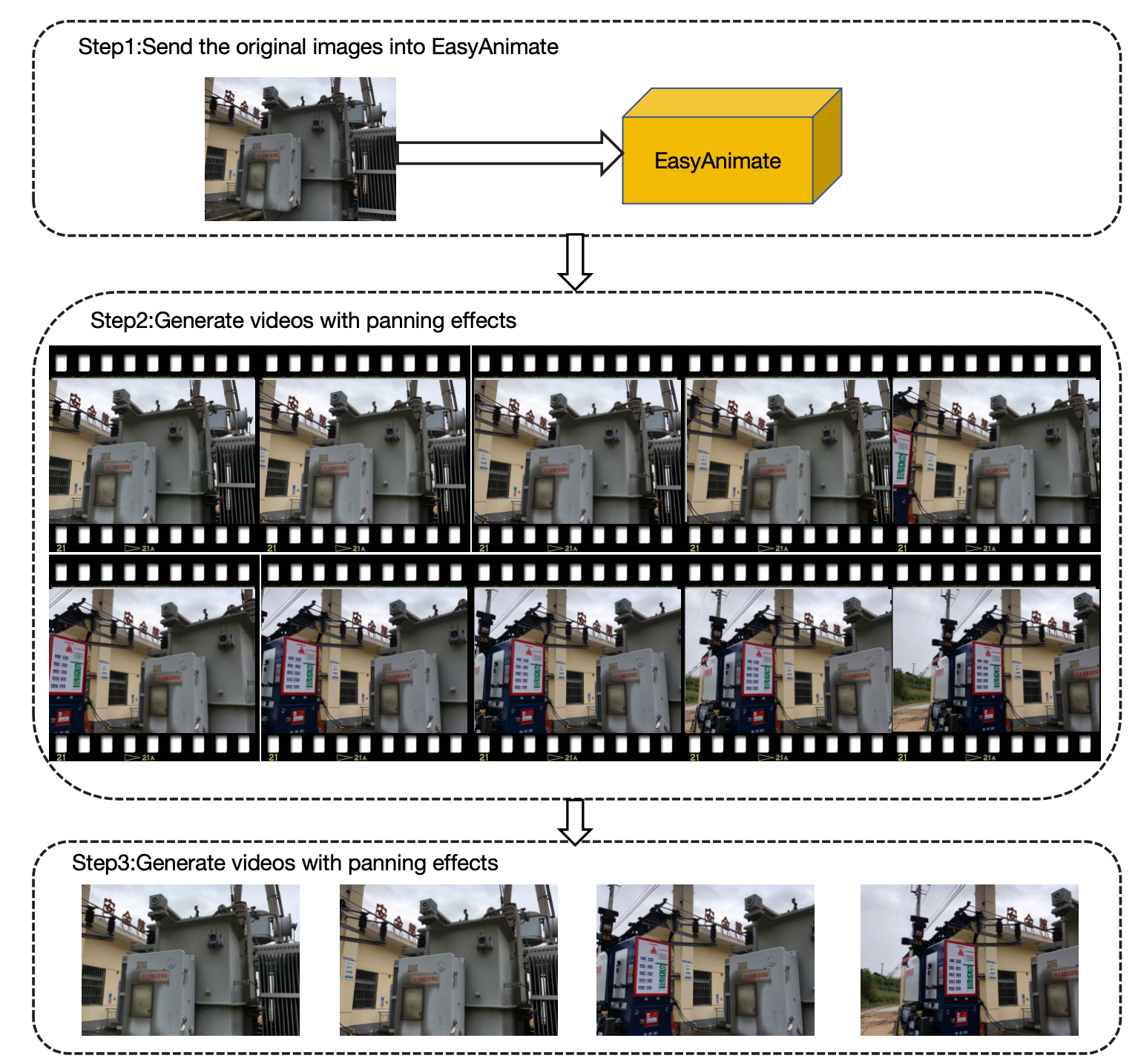}
    \caption{Fault Image Augmentation Diagram}
    \label{fig2}
\end{figure}

Based on the "Guidelines for the Assessment of the Condition of Power Transmission and Transformation Equipment"\cite{b22} by State Grid Corporation of China, we have developed corresponding prompts for different fault types. These prompts, along with defect images, were input into the GPT-4 model to generate fault analysis reports that include fault categories, causes, and repair recommendations. To ensure the accuracy and reliability of the reports, we enlisted five electrical experts to conduct a secondary review of the generated reports and corrected any unreasonable elements. The corresponding schematic diagram is shown in Figure \ref{fig1}.
Additionally, to expand the dataset, we used the image-to-video generation model EasyAnimate\cite{b23} to convert substation fault images into videos with panning effects. Each video lasts for 6 seconds, enhancing the diversity and applicability of the dataset.We captured screenshots every 1.5 seconds from each video to obtain images from different perspectives. This process increased the dataset to 40,000 images, as illustrated in Figure \ref{fig2}.

\section{Experimental Methodology}
The model’s responses sometimes lack electrical expertise and relevance. For example, in the analysis of a damaged transformer breather, it may respond simply with “Breather damage detected, please repair promptly,” lacking cause analysis and specific repair suggestions. By integrating a knowledge base for knowledge enhancement, the model can generate more professional and comprehensive analysis reports.For instance:
\begin{itemize}
    \item Fault Type: Damage to the oil-immersed transformer breather
    \item Fault Cause: Excessive internal pressure leading to aging or failure of the breather's sealing material
    \item Repair Recommendations: Check the breather for cracks or damage and replace it if necessary. Inspect the internal pressure of the transformer to ensure it is within normal ranges...
\end{itemize}

This enhancement method allows for a more detailed description of the fault type, specific causes, and repair suggestions, significantly improving the professionalism and practicality of the response.
\subsection{Enhancing fault analysis with knowledge}
\begin{figure}
    \centering
    \includegraphics[width=1\linewidth]{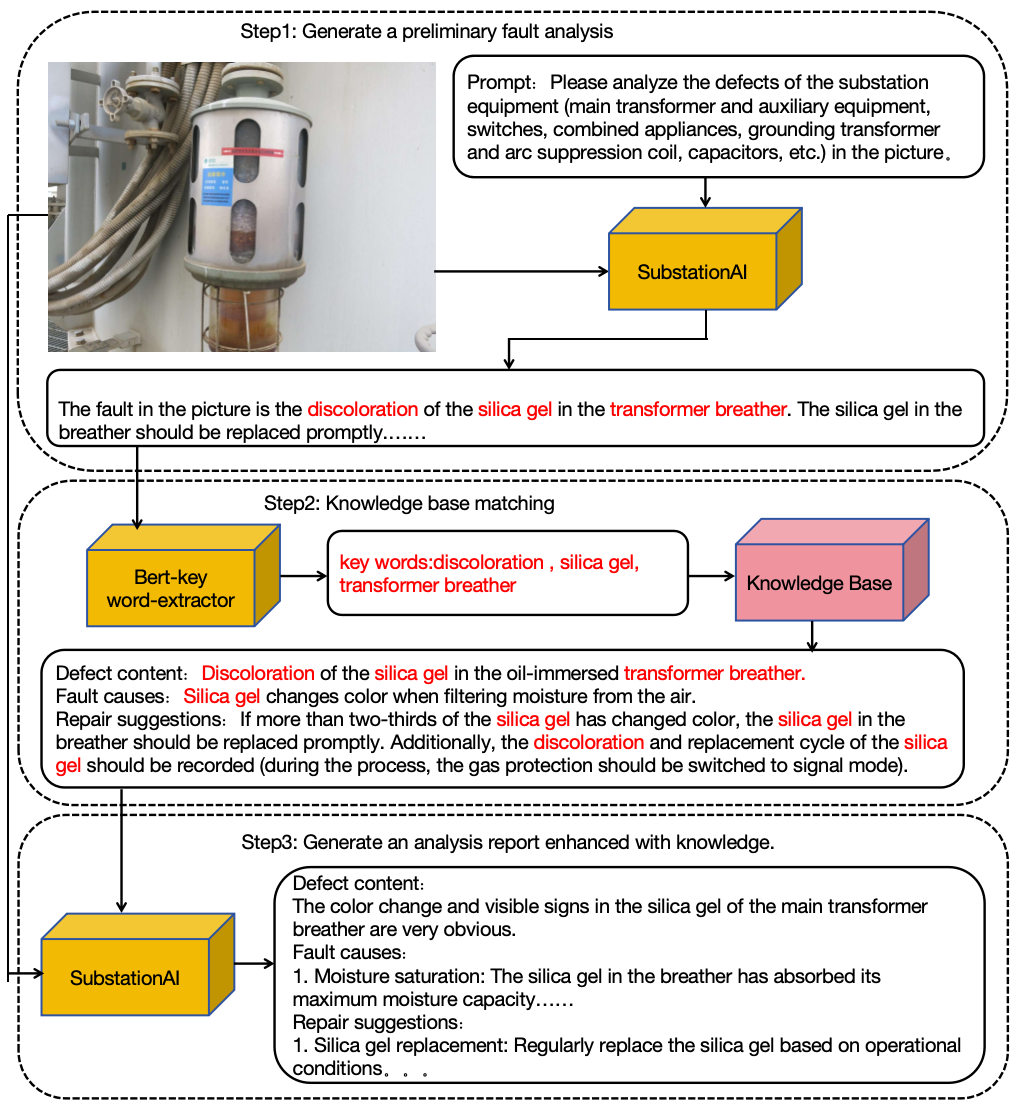}
    \caption{Knowledge Enhancement Diagram}
    \label{fig3}
\end{figure}

To improve the quality of fault analysis reports, we developed a specialized knowledge base for substation fault analysis based on the "Guidelines for Condition Assessment of Transmission and Substation Equipment" by the State Grid Corporation of China. We designed a knowledge enhancement method that starts with the model generating an initial fault description. BERT-based keyword extraction then identifies key information, and the system searches the knowledge base to retrieve relevant points. The highest-scoring knowledge points are selected and integrated into the model, enhancing the generated report's detail and accuracy. The workflow of this method is shown in Figure \ref{fig3}.

\section{Experiments}
\subsection{Experimental Setups}
We selected LLAVA1.5-7B as the base model for training SubstationAI. For this model, we conducted LoRA parameter fine-tuning on the constructed dataset over a period of 60 hours. The specific configuration included using three 3090 GPUs, with a learning rate set to 1e-4, training for 20 epochs, and a batch size of 10. The parameters for the LoRA layers were set with a rank of 64, an alpha value of 16, and a dropout rate of 0.05.
\subsection{Baselines}
To validate the performance of SubstationAI, we selected six baseline models for comparative analysis, including GPT-4, Claude-3\cite{b24}\cite{b25}\cite{b26}, LLAVA1.5-7B\cite{b27}\cite{b28}, VisualGLM-6b\cite{b29}\cite{b30}, Qwen2-VL-7B\cite{b31}\cite{b32}, and MiniGPT-4\cite{b33}\cite{b34}. A total of 1,000 samples were randomly selected from the dataset as benchmark data, covering 14 common types of substation faults. Subsequently, the six selected models were used to generate corresponding fault analysis reports. To evaluate the quality of these reports, we invited five engineers with a background in the electric power industry to rate the reports generated by each model based on predefined evaluation criteria. The detailed scoring results are presented in Table\ref{tab1}.
\begin{table}[htbp]
\caption{Model Score Comparison Analysis Table}
\begin{center}
\begin{tabular}{|c|c|c|c|c|c|}
\hline
\textbf{\textit{Model}} & \textbf{\textit{Acc.}} & \textbf{\textit{Cla.}} & \textbf{\textit{Com.}} & \textbf{\textit{Pra.}} & \textbf{\textit{Ave.}} \\
\hline
GPT-4 & 3.43 & 3.37 & 3.35 & 3.25 & 3.35 \\
\hline
Claude-3 & 3.30 & 3.46 & 3.50 & 3.23 & 3.38 \\
\hline
LLAVA1.5-7B & 2.94 & 3.01 & 3.11 & 2.87 & 2.98 \\
\hline
VisualGLM-6B& 2.83 & 3.04 & 2.91 & 2.83 & 2.90 \\
\hline
Qwen2-VL-7B& 2.79 & 2.76 & 3.03 & 2.86 & 2.86 \\
\hline
MiniGPT-4& 2.33 & 2.42 & 2.53 & 2.46 & 2.44 \\
\hline
SubstationAI& \textbf{4.32} & \textbf{4.11} & \textbf{4.41} & \textbf{4.27} &\textbf{ 4.28} \\
\hline
\end{tabular}
\label{tab1}
\end{center}
\end{table}

\subsection{Evaluation of Fault Analysis Report}
As shown in the results, SubstationAI achieved the highest scores across all evaluation metrics, significantly outperforming other models, including well-known multimodal models like GPT-4 and Claude-3. SubstationAI scored 4.32, 4.11, 4.41, and 4.27 in professional accuracy, clarity, completeness, and practicality, respectively, with an average score of 4.28. These results highlight its superior ability to generate high-quality fault analysis reports, with all metrics scoring above 4, validating its outstanding performance. In comparison, while GPT-4 and Claude-3 performed well on some metrics, their overall scores were slightly lower, emphasizing the benefits of our knowledge enhancement approach and customized model design. Additionally, LLAVA1.5-7B, the base model for SubstationAI, performed best among models of similar scale, demonstrating its strengths as a foundational model.
\subsection{Ablation experiment}
To validate the effectiveness of each method, we conducted ablation experiments, testing the use of only SFT, only knowledge enhancement, and replacing knowledge enhancement with Zero-shot-CoT\cite{b35}. The results\ref{tab2}, as shown in Table 2, indicate that SFT had the most significant impact on improving the model's performance across various metrics. Although the enhancement effect of knowledge enhancement alone was less pronounced than SFT, its combination with SFT led to further performance gains. In contrast, Zero-shot-CoT's performance was somewhat limited, likely due to the absence of professional knowledge prompts.
\begin{table}[htbp]
\caption{blation Experiment Score Comparison Table}
\begin{center}
\begin{tabular}{|c|c|c|c|c|c|}
\hline
\textbf{\textit{Model}} & \textbf{\textit{Acc.}} & \textbf{\textit{Cla.}} & \textbf{\textit{Com.}} & \textbf{\textit{Pra.}} & \textbf{\textit{Ave.}} \\
\hline
Original & 2.94 & 3.01 & 3.11 & 2.87 & 2.98 \\
\hline
SFT & 3.62 & 3.54 & 3.64 & 3.53& 3.58 \\
\hline
COT & 3.07 & 3.09 & 3.22 & 3.01 & 3.10 \\
\hline
KE. & 3.23 & 3.27 & 3.24 & 3.26 & 3.25 \\
\hline
SFT-COT & 3.71 & 3.62 & 3.74 & 3.66 & 3.68 \\
\hline
SFT-KE.& \textbf{4.32} & \textbf{4.11} & \textbf{4.41} & \textbf{4.27} & \textbf{4.28} \\
\hline
\end{tabular}
\label{tab2}
\end{center}
\end{table}
\subsection{Case Study}
\begin{figure}
    \centering
    \includegraphics[width=1\linewidth]{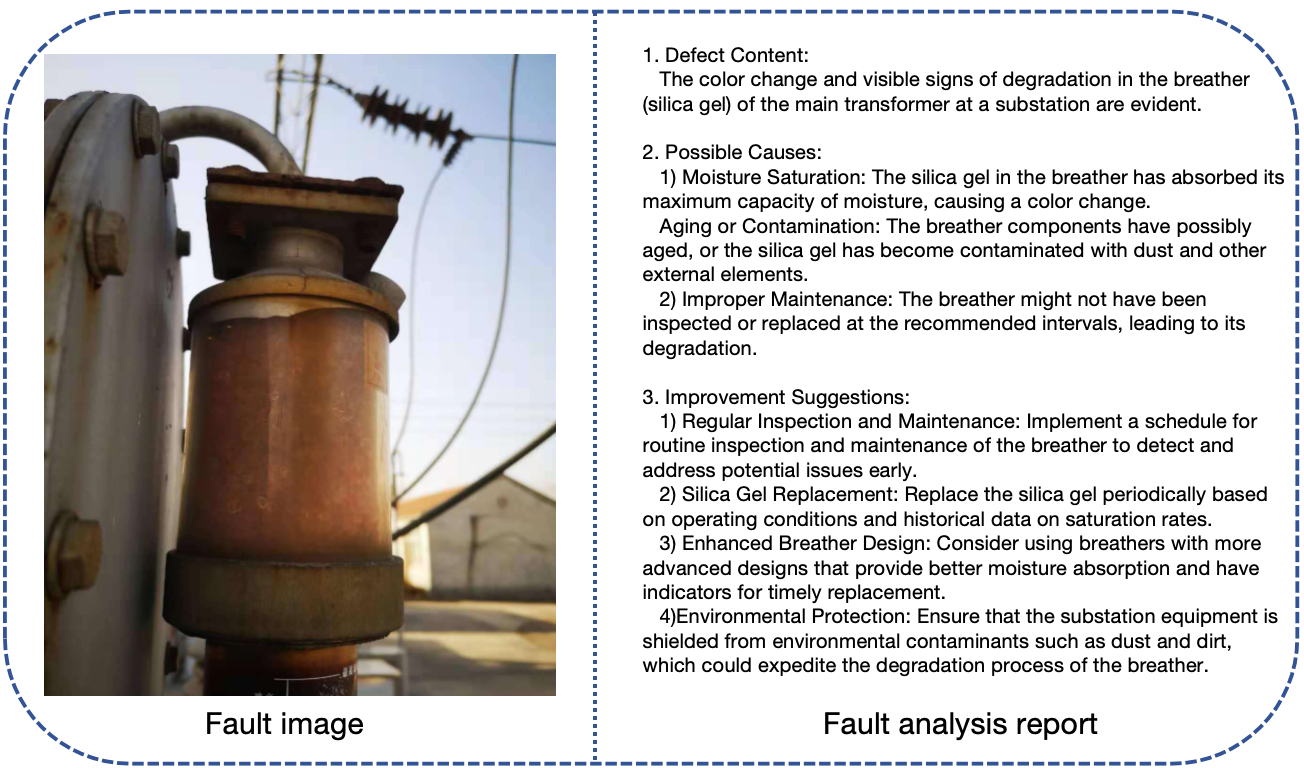}
    \caption{Fault Analysis Report Example}
    \label{fig4}
\end{figure}

This section presents the fault analysis report generated by SubstationAI, as shown in Figure\ref{fig4}. SubstationAI can accurately identify common equipment issues, such as color changes and visible signs in the main transformer's breather (silica gel). It not only provides a comprehensive analysis of the fault causes but also offers targeted improvement suggestions, effectively guiding on-site engineers in equipment maintenance and fault handling. By integrating specialized knowledge from the electric power industry, SubstationAI generates high-quality fault analysis reports, demonstrating its significant potential for practical applications.
\section{Conclusions}
We constructed the first dataset for substation fault analysis and developed specialized evaluation metrics tailored to the characteristics of this task. Based on this foundation, we trained a model specifically for substation fault analysis, SubstationAI. Additionally, we developed a dedicated substation fault diagnosis knowledge base and designed corresponding knowledge enhancement methods to help improve the quality of the reports generated by the model. Experimental results show that SubstationAI scores higher on various evaluation metrics, significantly outperforming other multimodal large models such as GPT-4.

\vspace{12pt}

\end{document}